\documentclass[10pt,twocolumn,letterpaper]{article}

\usepackage{cvpr}
\usepackage{times}
\usepackage{epsfig}
\usepackage{graphicx}
\usepackage{amsmath}
\usepackage{amssymb}
\usepackage{caption}
\usepackage{subcaption}
\usepackage{bm}
\usepackage{multirow}
\usepackage{array}
\usepackage{url}

\usepackage{booktabs}
\usepackage{makecell}
\usepackage{pifont}
\newcommand{\cmark}{\ding{51}}%
\newcommand{\xmark}{\ding{55}}

\usepackage{amsfonts}
\newcommand{\overbar}[1]{\mkern 1.5mu\overline{\mkern-1.5mu#1\mkern-1.5mu}\mkern 1.5mu}

\usepackage[pagebackref=true,breaklinks=true,letterpaper=true,colorlinks,bookmarks=false]{hyperref}

\cvprfinalcopy 

\def\cvprPaperID{xxxx} 

\ifcvprfinal\fi
\begin{document}

\title{Detail-aware Deep Clothing Animations Infused with Multi-source Attributes}

\author{Tianxing Li, Rui Shi and Takashi Kanai\\
The University of Tokyo, Japan\\
{\tt\small li-tianxing@hotmail.com}
}

\twocolumn[{%
\renewcommand\twocolumn[1][]{#1}%
\maketitle

\begin{center}
    \includegraphics[width=\textwidth]{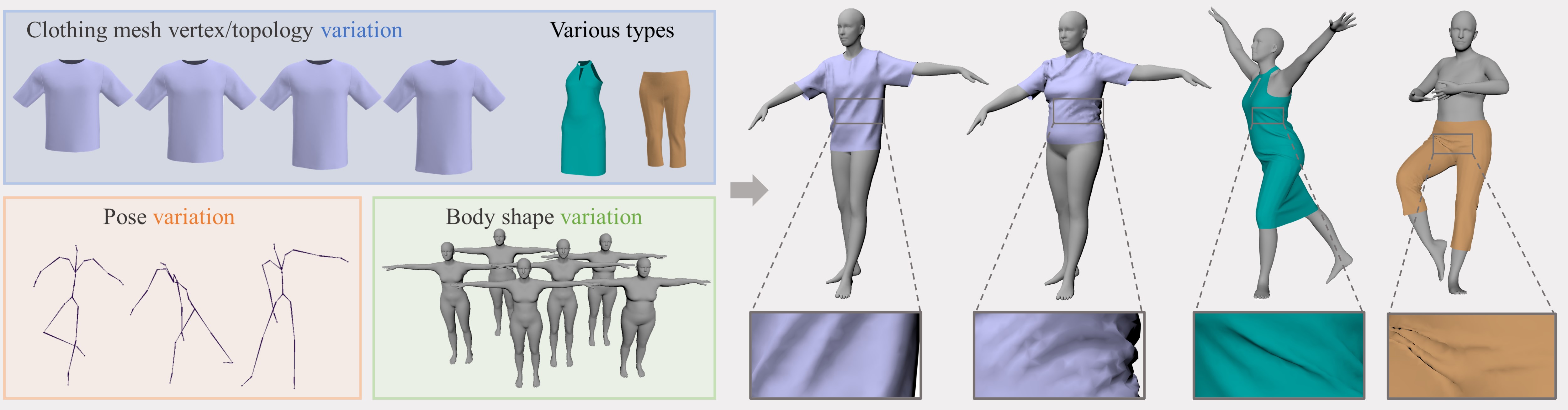}
    \captionof{figure}{We present a novel learning-based method for automatically generating detail-aware deformations for diverse garments worn by different body shapes in arbitrary poses. We train models based on clothing categories, and for each category, a limited number of models can predict individualized deformations caused by different attributes (clothing size, pose, and body shapes). }
    \label{fig:FigTeaser}
\end{center}%
}]


\begin{abstract}
  This paper presents a novel learning-based clothing deformation method to generate rich and reasonable detailed deformations for garments worn by bodies of various shapes in various animations. In contrast to existing learning-based methods, which require numerous trained models for different garment topologies or poses and are unable to easily realize rich details, we use a unified framework to produce high fidelity deformations efficiently and easily. To address the challenging issue of predicting deformations influenced by multi-source attributes, we propose three strategies from novel perspectives. Specifically, we first found that the fit between the garment and the body has an important impact on the degree of folds. We then designed an attribute parser to generate detail-aware encodings and infused them into the graph neural network, therefore enhancing the discrimination of details under diverse attributes. Furthermore, to achieve better convergence and avoid overly smooth deformations, we proposed output reconstruction to mitigate the complexity of the learning task.  Experiment results show that our proposed deformation method achieves better performance over existing methods in terms of generalization ability and quality of details.   
\end{abstract}

\section{Introduction}

Clothing animation is a fundamental topic in computer graphics, aiming to generate realistic clothing deformation effects for many applications including virtual try-on, video games, and films. With the progress of the graphics field, users are paying more attention to the visual effects of garments, such as how can the garment and body be made to interact more realistically, and how the wrinkles increase or decrease with  different movements. High-quality clothing deformations provide users with better experience during online shopping for example or help stimulate interest in continuous entertainment.

To meet the needs of producing high-quality clothing animations, predominant approaches are based on physics-based simulation \cite{Narain12, nealen2006physically}. Despite the convincing effects provided by these methods, deployment to real-time applications is still challenging due to the high costs of computer simulation process and the sensitivity of results to parameter settings.

To overcome high computational costs and simplify the deformation process, learning-based solutions \cite{clothing10} are proposed to approximate clothing deformations according to relevant influencing factors (\textit{e.g.}, motion and shape of the body).  While these methods can roughly imitate the behavior of clothing animation, there still remain issues in terms of generalization and quality of details.

Most state-of-art studies \cite{santesteban2019virtualtryon, patel20tailornet, tiwari20sizer} adopt multilayer-perceptron (MLP) models to predict the nonlinear deformations of garments. Although the predicted results contain plausible wrinkles, due to the fixed dimensions of input and output vectors (which are usually related to the number of vertices), training and testing targets are enforced to have the same number of vertices, so the trained model is unable to generalize to new garments with new mesh topologies. Furthermore, because of the limited ability of MLPs to understand 3D information, a great number of parameters is usually required to realize the deformation approximation for  specific mesh topologies. On the other hand, solutions based on graph neural networks \cite{20Chentanez, GarNetPlus20} can effectively address the generalization limitation of MLPs, however, the approximated garments tend to be overly smooth and lack rich wrinkles \cite{GarNet19}. To enable realistic clothing deformations, existing graph learning-based research \cite{vidaurre2020virtualtryon} has to trade pose-variation for realism, which only predicts the deformation in t-pose. 

\begin{figure*}[t]
\includegraphics[width=1\textwidth]{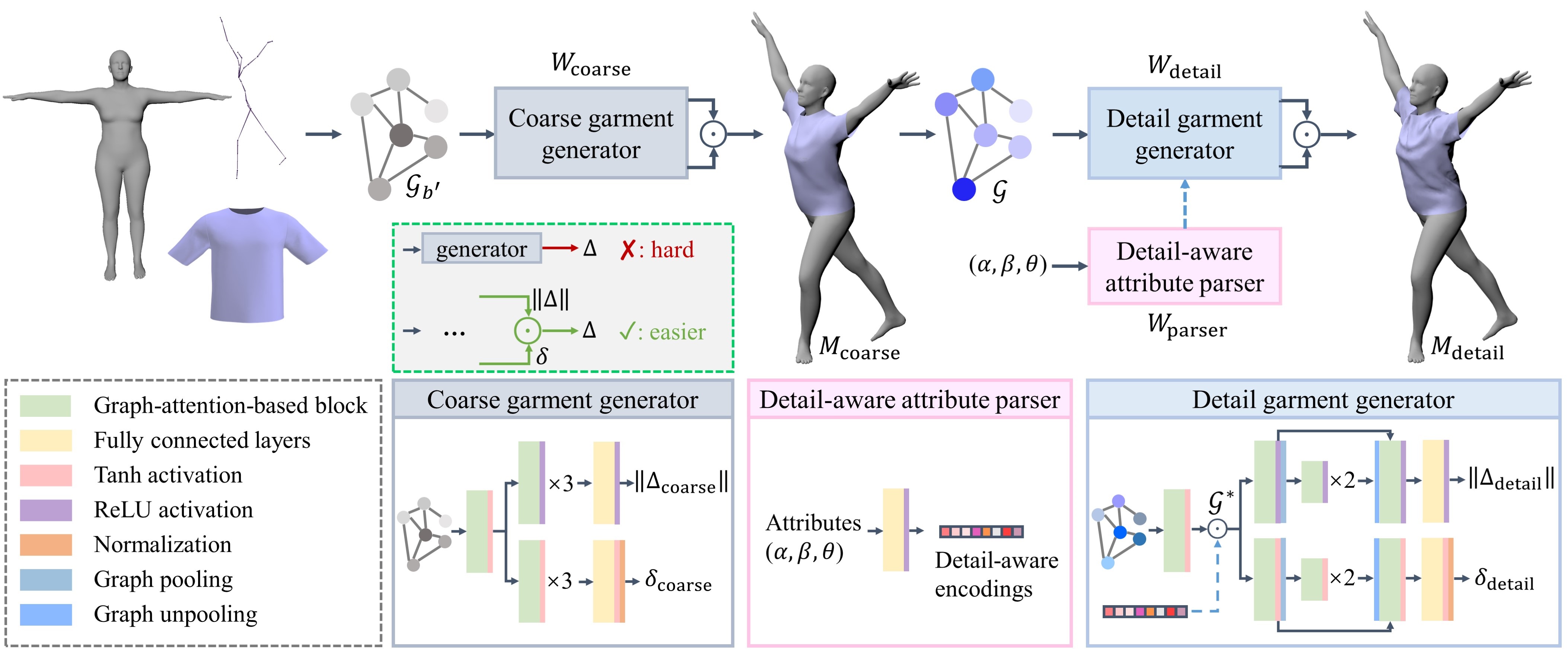}
\centering
\caption{Overview of proposed method pipeline. Given a garment with an arbitrary mesh topology, a target body with any shape, and a random animated posture, our method is able to approximate high-quality clothing deformation with expressive detail wrinkles. Our key contribution is to address the challenge of ``complexity'' by designing a two-step framework with ideas of proposing the fit parameter $\alpha$, detail-aware attribute parser, and output reconstruction. First, the constructed graph $\mathcal{G}_{b'}$ is fed into a coarse garment generator $W_{\text{coarse}}$ to predict the decomposed components $\|\Delta_{\text{coarse}}\|$ and $\delta_{\text{coarse}}$ of the coarse corrective displacement therefore realizing coarse deformation prediction $M_{\text{coarse}}$. Next, we build a graph $\mathcal{G}$ based on the generated deformation $M_{\text{coarse}}$. Instead of directly applying attributes to each graph node, we further propose a detail-aware attribute parser $W_{\text{parser}}$ to generate detail-aware encodings and infuse them into the original graph to obtain the representative $\mathcal{G}^*$. Then, a detail garment generator is designed to process features of $\mathcal{G}^*$ and output $\|\Delta_{\text{detail}}\|$ and $\delta_{\text{detail}}$ in each branch. Two predictions are finally multiplied and added to the $M_{\text{coarse}}$ to realize the ultimate detail clothing deformation $M_{\text{detail}}$.
}  
\label{fig:FigPipeline}
\end{figure*}

The main reason why learning-based methods for clothing animation need to weigh the above aspects is: the extreme complexity of the fine deformation prediction of garments in multiple states (under various postures, worn by various bodies, \textit{etc.}). Our method essentially overcomes this ``complexity'' and uses fewer models to efficiently generate high-quality deformations with fine details. Deformations can be approximated in two steps (Figure \ref{fig:FigPipeline}): 1) learn a model to globally drape the garment on the target body in a certain pose, 2) learn an additional model to produce the high-frequency wrinkles based on the corresponding coarse deformation. Specifically, our technical contributions are three-fold: 

\begin{itemize}
\item To account for complicated and irregular detailed wrinkles, we first discuss that the fit between the garment and body influences the degree of wrinkles: loose clothes have smoother, sparser, and wide wrinkles, while tight clothes have thinner, denser, and narrow wrinkles. Therefore, we parametrize the relationship and propose the fit parameter, which is regarded as one of the attributes.
 
\item To make the model generalized and effectively map relevant influencing attributes (\textit{i.e.}, fit, body shape, and pose) to deformation details, we design an attribute parser to generate detail-aware encodings and then infuse them into the graph neural network. This infusion maps the original graph features to representative features that are adaptive to the corresponding attributes, providing a meaningful signal to the model and learning realistic deformations in a detail-aware manner.  

\item To facilitate the deformation learning and achieve high-quality predictions, we address  complexity fundamentally from the novel perspective of output reconstruction. Existing studies always directly output the three-dimensional vector (position or displacement) of each vertex where the value of each dimension ranges from negative infinity to positive infinity, which makes it difficult for the training to converge to a reasonable range and the prediction results tend to be overly smooth. To address this problem, we decompose the output vector as the combination of magnitude and direction where the value range of the magnitude is greater than zero and the value range of the direction is from -1 to 1. This strategy plays a crucial role in the learning of fine deformations, since it greatly reduces the range of output variables, thereby mitigating the complexity of the task.  
\end{itemize}

To the best of our knowledge, our study has been the first to enable unified models to realize detail-aware deformations for garments with various mesh geometries worn by diverse body shapes in any posture. Our experiments confirm that our proposed method outperforms existing clothing animation methods in terms of generalization and deformation quality.

   
\section{Related Work}
In this section, we first discuss existing clothing animation methods by classifying them into physics-based simulation and learning-based models. Then, we also introduce the latest investigations on learning-based deformation.    

\noindent \textbf{Physics-based simulation.} Pioneering studies achieve realistic clothing animations based on geometric constraints \cite{Realtime04, Selle2009RobustHC, Rohmer10}, however, they always suffer from instability and high computational cost. In order to make the simulation efficient, research in \cite{Muller10} computes wrinkles by a static solver and adds them on the coarse base mesh. As a similar idea on adding fine details on low-quality cloth, Gillette \textit{et al.} \cite{Gillette15} propose tracing wrinkle paths on the coarse mesh following the per-triangle compression field. To accelerate the computation, recent researchers are also making efforts to improve GPU-based algorithms. For example, yarn-level contact can be modelled implicitly with GPU in \cite{Yarn14}. Ni \textit{et al.} \cite{ni2015scalable} present an algorithm to simulate cloth with complex collisions using a parallel run-time system. To exploit high parallel performance, a matrix assembly algorithm is proposed \cite{cama16} which can accurately solve the linear system. Despite the realistic results and improved run-time of physics-based simulation, professional parameter setting and potential instability pose as obstacles to achieving reduction of computational cost of physics-based simulation, and nonlinear behaviors like the physics-based simulation.    

\noindent \textbf{Learning-based clothing models.}
Inspired by the success of deep learning, a number of works are attempting to learn the deformation as a function of relative parameters.

To resolve the high computational costs of physics-based simulation while realizing nonlinear clothing behaviors, Santesteban \textit{et al.} \cite{santesteban2019virtualtryon} propose a two-level strategy to generate clothing deformations, where the first step is to use MLPs to learn the global fit and the second step is to use recurrent neural networks to learn the wrinkles. Also in order to estimate cloth deformations with fine details, TailorNet \cite{patel20tailornet} adopts multiple MLPs to realize the task, in which low-frequency deformations are predicted using a simple MLP model, and high-frequency deformations are predicted using the mixture of multiple MLPs. To model how people wear the same garments in different sizes, Tiwari \textit{et al.} \cite{tiwari20sizer} propose a SizerNet to approximate the wearing effect of a garment in different sizes. Because the dataset only consists of A-pose garments and garments, the proposed method cannot generate a variety of deformations in different poses. To solve the garment-body interpenetration, novel garment space is proposed in \cite{santesteban2021self}, which eliminates the need for any postprocessing steps. Although these studies have achieved success in the automatic clothing deformation approximation with fully-connected layers, a common limitation of these methods is the generalization ability, \textit{i.e.}, independent training is always required when deforming new garments with new mesh topologies.

To address the fundamental limitation of generalization in learning-based deformations, research tries to approximate the clothing deformation using graph neural networks which can  handle 3D data in non-Euclidian domains. The latest study in \cite{20Chentanez} introduces a graph neural network with a novel convolution operator for cloth and body skin deformation approximation. The proposed solution is specifically for triangle meshes. Inspired by point cloud processing, Gundogdu \textit{et al.} \cite{GarNet19}  introduce the framework based on PointNet for clothing animation. The results look plausible  but sometimes tend to be overly smooth. Focusing on fast clothing deformation, Vidaurre \textit{et al.} \cite{vidaurre2020virtualtryon}  present a fully convolutional graph neural network (FCGNN) to predict deformations with fine-scale details. The framework consists of two graph neural networks with the same structure and a different number of layers, which respectively predict the coarse draping and refinement. The proposed pipeline can generalize to unseen mesh topologies, garment parameters, and body shapes. However, the prediction is only for one pose and does not consider pose variations.

Alternatively, some studies achieve clothing animation from the perspective of computer vision. Research in DeepWrinkles \cite{Lahner_2018_ECCV} learns a conditional adversarial network to generate high-frequency details in normal maps. Recently, Zhang \textit{et al.} \cite{wrinkle21} tackle the generalization problem and make it possible to transfer details across the normal maps of different garments. Realistic clothing animation can also be achieved by the deforming garment with the displacement map \cite{pixel20}. However, it falls short when applied with loose garments for the body.  

\noindent \textbf{Learning-based deformation.}
Several methods also apply data-driven models to deformation approximation for animated characters. Loper \textit{et al.} \cite{smpl15} present a learned skinned multi-person linear model (SMPL) of human body shape and pose-dependent shape variation. Based on this, dynamic blend shapes are predicted in \cite{18Casas, Santesteban2020SoftSMPLDM} to enrich soft-tissue effects. These approaches can be well generalized to new shapes and motions, but only work for body meshes with a fixed number of vertices. To make film-quality characters run in real-time, Bailey \textit{et al.} \cite{bailey2018fast}  train multiple MLPs for one specific character. The generalization problem is solved in \cite{liu2019neuroskinning}, which uses graph neural networks to predict skinning weights for game characters with complicated dressing. Research in \cite{RigNet} also utilizes the graph-learning-based methods to predict the number of joints and skinning weights. To achieve realistic deformation, the nonlinear corrections is predicted in each pose step \cite{DenseGATs20, li2021MultiResGNet} by using the improved graph neural networks. Inspired by these methods, in this work, we adopt the SMPL model as the base body and design graph- learning-based models to achieve the clothing deformation with good generalization ability and high-quality results.
\section{Overview}
Given a garment with arbitrary mesh topology, a target human body with any shape, and a series of poses in motion, our goal is to automatically generate realistic clothing deformation with fine-scale wrinkles. Training and predicting this task are not simple due to the extremely changeable deformation details. To address this challenge, we first propose a fit attribute that can affect the details of wrinkles to a large extent (Section \ref{sec:fit}). Together with shape and pose attributes, the multi-source attributes enable us to predict more realistic clothing deformations and can give the model good generalization capabilities. Next, to fundamentally mitigate the complexity of the task, while ensuring high-quality deformation effects, we propose a new perspective of output reconstruction (Section \ref{sec:output}). Unlike the direct prediction of the displacement deviation of each vertex in all previous studies, we decompose this deviation so that the numerical range of the prediction target is greatly reduced. With these strategies, we introduce a pipeline that divides the deformation into two steps. The first step (Section \ref{sec:coarse}) is to learn a coarse garment generator to globally produce smooth clothing deformations with global draping effects. As depicted in Figure \ref{fig:FigPipeline}, we use a coarse garment generator $W_{\text{coarse}}$ to achieve this, where  $W_{\text{coarse}}$ is designed with two branches consisting of graph-attention-based blocks and fully connected layers. Next, the second step (Section \ref{sec:fine}) is to further enhance details based on the coarse garment. Because of the complexity of this step, as shown in Figure \ref{fig:FigPipeline}, we design an attribute parser $W_{\text{parser}}$ to generate detail-aware encodings based on multi-source attributes and then infuse them into  detail garment generator $W_{\text{detail}}$ to  generate rich and plausible wrinkles locally. With the help of $W_{\text{parser}}$, excessive smoothness can be avoided in deformations generated by  $W_{\text{detail}}$ to a certain extent.   

\section{Approach}
\subsection{Attributes Affecting Deformation} \label{sec:fit}
\begin{figure}[t!]
  \centering
  \includegraphics[width= 1\linewidth]{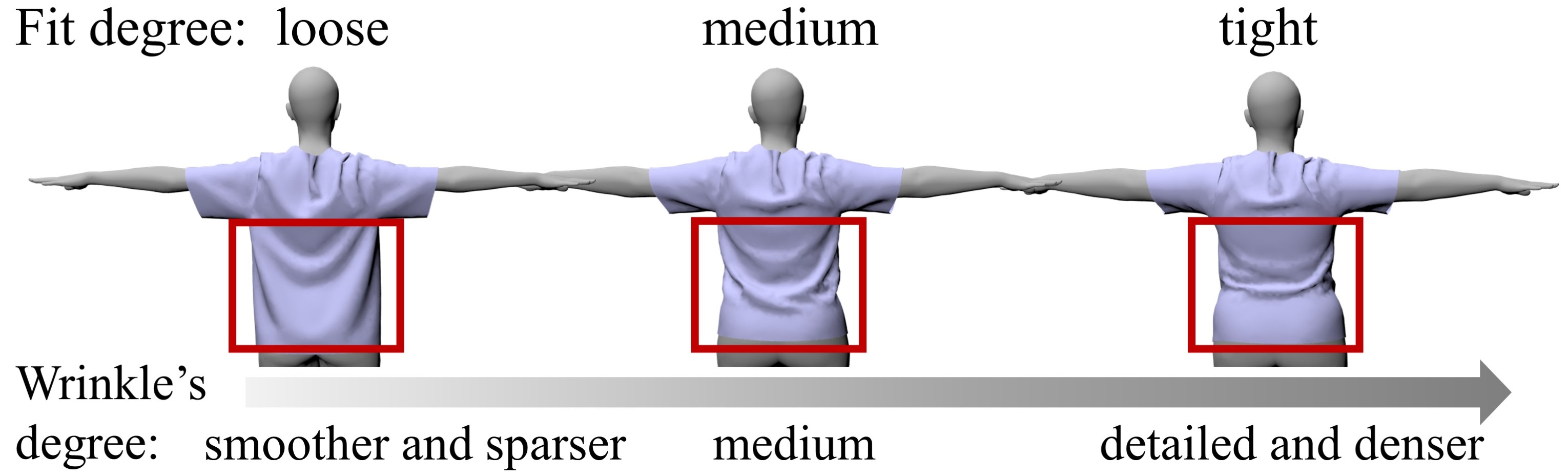}
  \caption{\label{fig:FigFitPara} Fit of garment and body influences wrinkles.
}
\end{figure}

To achieve complex clothing deformations, we first observed parameters that affect the quality of deformations. In the real scene, the different relationship between the clothes and the body, \textit{i.e.}, the different degree of fit, the deformation presented is different in terms of the whole (global) and detailed (local) effects. As shown in Figure \ref{fig:FigFitPara}, for the fixed material, when the fit degree is from loose to tight, the wrinkles of garments are from smoother and sparser (with a wider wrinkle width) to finer and denser (with a narrower wrinkle width). This observation indicates the need to generate fit parameter as one of the network inputs, which can better target the different suitability of clothing, thereby producing more realistic deformations. Next, we will describe how to build this relationship between garment and body and how to express this variation.

For the target body, we adopt the SMPL \cite{smpl15} model which represents the human body $M_{b}$ parameterized by shape ($\beta$) and pose ($\theta$):
\begin{align}
M_{b} &= W_{\text {smpl}}(\overbar{M}_{b}(\beta, \theta), J(\beta), \theta, \boldsymbol{\mathcal{W}}),\\
\overbar{M}_{b} &= T + B_s(\beta) +B_p(\theta), 
\end{align}
where the learned skinning function  $W_{\text {smpl}}(\cdot)$ is applied to deform the unposed body $\overbar{M}_{b}(\beta, \theta)$ with skinning weights $\boldsymbol{\mathcal{W}}$ of the skeleton $J(\beta)$. The unposed $\overbar{M}_{b}(\cdot)\in \mathbb{R}^{N_b\times 3}$ is computed by applying  blend shapes $B_s(\beta)$ and $B_p(\theta)$ on the based mesh $T \in  \mathbb{R}^{N_b\times 3}$, \textit{i.e.}, adding corrective displacements per vertex. These blend shapes are respectively dependent on shape $\beta \in  \mathbb{R}^{|\beta|} $ and pose $\theta \in  \mathbb{R}^{|\theta|} $ where $|\beta|$ and $|\theta|$ are the coefficients of shape and pose. 

Given $N_{\text {pair}}$ garment-body pairs composed of different garment lengths (with different topologies) and different SMPL body shapes, we first select evenly distributed vertices on the unposed body and calculate the distance between these vertices to the garment:  
\begin{equation}
\textbf{D} = \mathop{\big|\big|} \limits^{N_{\text{pair}}} \|\overbar{M}_{g^*} - \textbf I^*\overbar{M}_{b}\|,
\end{equation}
where $\textbf I^* \in \{0,1\}^{N_{g^*} \times N_b}$ is the indicator matrix to evaluates if the garment vertex is associated with the body vertex. The number of selected vertices is $N_{g^*}$, which is equal to the minimum number of vertices in all garments. $\overbar{M}_{g^*} \in \mathbb{R}^{N_{g^*} \times 3}$ is the part of or all of the $\overbar{M}_{g}$. $\| \cdot \|$ denotes the Euclidean norm operation. $\big|\big|$ means the concatenation of $N_{\text {pair}}$ vectors. Therefore, $\textbf{D} \in \mathbb{R}^{N_{g^*}\times N_{\text {pair}}}$ stores all the distance information of each vertex of each pair. Then, we seek an expression to represent this information concisely and intuitively. 

We compute the fit parameter using factor analysis to model the variance along each vertex independently. Considering the speed of convergence, we apply a faster method named SVD-based likelihood optimization \cite{SHI202075}. Factor analysis in matrix term is defined as:
\begin{equation}
\textbf{D} - \mu \approx \textbf{L} \textbf{A},
\end{equation}
where $\mu \in \mathbb{R}^{N_{g^*}}$ is the mean vector. $\textbf{L} \in \mathbb{R}^{N_{g^*} \times F}$ and $\textbf{A} \in \mathbb{R}^{F \times N_{\text {pair}}}$ denote the loading matrix and factors.  In this way, $\textbf{A}$ consists of $N_{\text {pair}}$ of vector $\alpha = [a_1, a_2,...,a_F] \in \mathbb{R}^{F}$ which provides an $F$-dimensional suitableness representation for each garment-body pair. We call this parameter $\alpha$ as the fit attribute. In addition, for the fine clothing deformation, body shape and pose also have an impact on the detail folds. Hence, we refer to these three parameters $(\alpha, \beta, \theta)$ collectively as multi-source attributes. These multi-source attributes play a key role in generating detailed deformations, which are taken as the input of $W_{\text{parser}}$ introduced in Section \ref{sec:fine} .  


\subsection{Mitigating Complexity of Learning} \label{sec:output}
Most deformation approximation studies are plagued by the problem of highly nonlinear output. Since the value of three-dimensional vertices offsets adjustment ranges from negative infinity to positive infinity, several researches
have been able to only leverage a large number of fully connected layers while sacrificing generalization \cite{patel20tailornet}, or only perform prediction for one pose \cite{vidaurre2020virtualtryon} to ensure quality. So far, there has been no research attempting to solve the problem fundamentally from the perspective of reconstructing output.

In our work, we propose an output reconstruction method by decomposing the output vector of each vertex into the magnitude and direction:
\begin{equation}
\Delta_i = \|\Delta_i\| \odot \delta_i,
\end{equation}
where the original output is $\Delta_i \in \mathbb{R}^3$, the decomposed magnitude is $\|\Delta_i\| \in \mathbb{R^+}$, and the direction is $\delta_i\in \mathbb{R}^3$. Unlike other learning-based methods which directly predict $\Delta_i$ with a wide value range of ($-\infty, +\infty$), our method indirectly predicts the vector's magnitude $\|\Delta_i\|$ and the direction $\delta_i$ with the narrow value range of ($0, +\infty$) and ($-1, 1$). In our two generators (shown in Figure \ref{fig:FigPipeline}), both networks are designed with two branches in order to predict the decomposed items separately. In addition, based on the value characteristics, we adopt different activation functions in two branches: ReLU is used in the $\|\Delta_i\|$ branch to output positive values; Tanh is used in the $\delta_i$ branch to map the resulting values between -1 to 1. Thus, in contrast to the original output $\Delta_i$  with the infinite degree of freedom, the value range of our decomposed output is greatly ``narrowed'', and with the help of the activation function, it can be ensured that the output is always within a reasonable range. 

With the two approximated items of $\|\Delta_i\|$ and $\delta_i$, we finally multiply them together to obtain the final nonlinear offset vector. The decomposition step does not seem complicated, and it plays a crucial role that greatly mitigates the complexity of learning and can generate better quantitative and qualitative results. 

\subsection{Coarse Garment Prediction} \label{sec:coarse}
As stated in previous work \cite{santesteban2019virtualtryon, patel20tailornet, vidaurre2020virtualtryon}, directly regressing clothing deformations as a function of designed parameters with one model will result in unrealistic results. Therefore, the final deformation process must be divided into several steps to perform approximations. In this work, we also decompose clothing deformation into coarse deformations with the overall fit into and detailed deformations with fine-scale wrinkles. 

The goal in the first step is to achieve plausible clothing coarse deformation $M_{\text{coarse}}$ in the garment worn by the target body $M_b$ in a certain animated pose: 
\begin{equation}
M_{\text{coarse}} = \textbf I M_b + \Delta_{\text{coarse}},
\end{equation}
where $\textbf I \in \{0,1\}^{N_g \times N_b}$ refers to the indicator matrix of the association between garment and body vertices. 
For the remaining residual part $\Delta_{\text{coarse}} \in \mathbb{R}^{3 \times N_g}$, we aim to learn a model $W_{\text{coarse}} $ to automatically infer the offsets.

With the garment and animated body, we first need to construct a parametric space that can concisely express useful information for coarse deformation without ignoring the spatial information. Therefore, we consider the input of our network to be a mesh graph: $\boldsymbol{\mathcal{G}}_{b'} = (\mathcal{V}_{b'},\mathcal{E}_{b'},\mathbf{U}_{b'})$ which stores features of a part of body vertices $\mathcal{V}_{b'} = \left\{1, ..., N_{b'}\right\}$ ($N_{b'}$ is a part of $N_b$), edges $\mathcal{E}_{b'} \subseteq \mathcal{V}_{b'}\times \mathcal{V}_{b'}$, and the adjacency representation $\mathbf{U} \in [0,1]^{N_{b'}\times N_{b'}}$. In particular, through the indicator matrix $\textbf I$, we map the $N_{g}$ vertices of the garment to the vertices $N_{b'}$ of the body one by one ($N_{b'} = N_{g}$), and also project the connectivity of these vertices to obtain edges $\mathcal{E}_{b'}$. Next, for each node $i \in \mathcal{V}_{b'}$, we need to assign attributes to make the node informative. Specifically, to encode the body mesh appearance, we append the vertex normal $n_{b'i} \in \mathbb{R}^{3}$ to each graph node; to reflect the body skinning features in different poses, we append $p_{b'i} \in \mathbb{R}^{3}$ to each node features, where the translation information is removed from the body vertex position to achieve translation-invariance. Moreover, to represent the relationship between the garment and body, we also attach the fit attribute to each node. Here, for simplicity and conciseness, the first  component $a_1$ of the fit attribute $\alpha$, which is the most discriminative one, is adopted. In total, each node $i$ in the graph $\boldsymbol{\mathcal{G}}_{b'}$ consists of three attributes and can be represented as: $v_{b'i} = [n_{b'i}, p_{b'i}, a_1] \in \mathbb{R}^{7}$.

Having the graph with defined features as input, next we need to design a model $W_{\text{coarse}}$ for acquiring the latent representation of the graph data and mapping it to the final prediction $\Delta_{\text{coarse}}$. To accomplish this task, there are two requirements for the design model $W_{\text{coarse}}$. Specifically, first, the model should have the generalization ability that is able to deal with garments with arbitrary mesh topologies. Second, the model should be able to infer the overall deformation of garments under various body shapes and postures according to the knowledge learned in the training process. To satisfy these needs at the same time, we adopt graph-attention-based (GAT) blocks which extend the original GAT structure \cite{velivckovic2017graph} with the self-reinforced stream \cite{DenseGATs20} for efficiently handling complicated 3D mesh features. By aggregating node features from the neighborhoods and strengthening self-features, such graph-attention-based blocks allow for acquiring the latent representations of irregular mesh graph  data without the need of knowing the graph structure upfront. 

As shown in in Figure \ref{fig:FigPipeline}, for coarse deformation prediction, first we apply one block in the first layer for dealing with the input graph, and then apply three blocks to each branch, \textit{i.e.}, the magnitude prediction branch and the direction prediction branch. The reason for designing two branches is that the value range of two predictions (as stated in Section \ref{sec:output}) is different and each branch needs to adopt a different activation function to ensure the range of the output value. In the last layer of two branches, linear transformation and corresponding activation and normalization are used therefore achieving the final predictions: the magnitude $\|\Delta_{\text{coarse}}\|$ and the direction $\delta_{\text{coarse}}$. The whole progress through the coarse generator can be expressed as: 
\begin{equation}
\|\Delta_{\text{coarse}}\|, \delta_{\text{coarse}} = W_{\text{coarse}}(\boldsymbol{\mathcal{G}}_{b'}).
\end{equation}
Lastly, two predictions are multiplied together to get the displacement $\Delta_{\text{coarse}}$ to the body. During the training, we minimize the MSE loss between the predicted displacement $\Delta_{\text{coarse}}$ and the ground truth $\Delta_{\text{coarse}}^{\text{GT}}$.

\subsection{Detail Garment Prediction} \label{sec:fine}
After obtaining the coarse clothing deformation, the next step is to realize detailed deformation with fine-scale wrinkles. Compared with coarse deformation that is easy to generate, detailed deformation is extremely difficult to obtain due to its complexity and volatility under various states. Despite research advances, existing learning-based studies always have to face the trade-off between the generalization ability of models and the fidelity of results, \textit{i.e.}, the model is only worked for fixed mesh topology \cite{patel20tailornet} or for fixed pose \cite{vidaurre2020virtualtryon, tiwari20sizer} and tends to produce overly smooth deformations \cite{GarNet19}. Even though significant efforts have been made on many aspects such as input improvement, network structure improvement, convolution operator change, and increase in the number of models, different degrees of wrinkles in diverse poses and shapes still cannot be stably learned and approximated.

To address these challenges, we propose the novel detail-aware attribute parser $W_{\text{parser}}$ and detail garment generator $W_{\text{detail}}$, where the key idea is to adjust the wrinkle-related adaptive distribution of the graph and transfer it through two branches for detailed deformation approximation.

On one hand, given the generated coarse deformation, we build a graph $\boldsymbol{\mathcal{G}} = (\mathcal{V},\mathcal{E},\mathbf{U})$, in which $\mathcal{V} = \left\{1, ..., N_{g}\right\}$ indicates clothing mesh nodes, $\mathcal{E}$ is mesh edges, and $\mathbf{U}$ is the adjacency matrix. For each node, the features are defined as: $v_{i} = [n_i, p_i, x_i]$, which consists of the vertex normal $n_i \in \mathbb{R}^{3}$, the vertex position $p_i \in \mathbb{R}^{3}$ (translation removed as stated in section \ref{sec:coarse}), and the distance to all joints $x_i \in \mathbb{R}^{S}$ ($S$ is the joint number).    

On the other hand, given a series of attributes that affect the degree of wrinkles, directly constructing graphs by assigning attributes (\textit{e.g.}, shape, pose, \textit{etc.}) to every single node and then forwarding them into the network is the most common strategy in previous graph-learning-based methods. However, it will lead to feature redundancy because attributes are independent on a single node. Therefore, we design a detail-aware attribute parser (as shown in Figure \ref{fig:FigPipeline}) that takes the multi-source attributes $(\alpha, \beta, \theta)$ as input and outputs the detail-aware encodings $W_{\text{parser}} (\alpha, \beta, \theta)$ that can adaptively adjust the graph feature distribution based on a given input instance. Specifically, detail-aware encodings are vectors where their dimensions equal to $d^{[1]}$, \textit{i.e.}, the dimension of the graph feature of each vertex after the first layer. Then, we element-wisely multiply it with the transformed graph along the feature dimension: 
\begin{equation}
\boldsymbol{\mathcal{G}}^{*} = W_{\text{parser}} (\alpha, \beta, \theta) \odot W_{\text{detail}}^{[1]}(\boldsymbol{\mathcal{G}}),
\end{equation}
where $\boldsymbol{\mathcal{G}}^{*}$  refers to the graph features infused the features after the first layer of the graph $W_{\text{detail}}^{[1]}(\boldsymbol{\mathcal{G}})$ with the detail-aware encodings  $W_{\text{parser}} (\alpha, \beta, \theta)$. In other words, the obtained features in a graph $\boldsymbol{\mathcal{G}}^{*}$ have been adaptively modified by high-dimensional attribute encodings, so that features can be expressed in a more detail-aware manner and be prepared for accurate prediction. 

We input the new graph $\boldsymbol{\mathcal{G}}^{*}$ into the following layers of $W_{\text{detail}}$ (except for the first layer $W_{\text{detail}}^{[1]}$, the remaining part can be expressed as $W_{\text{detail}}^{[2\text{-}L]}$ ). Similar to the coarse generator $W_{\text{coarse}}$, the detail generator $W_{\text{detail}}$ also has two branches, which respectively approximate decomposed detail output elements: the magnitude $\|\Delta_{\text{detail}}\|$ and the direction $\delta_{\text{detail}}$. Due to the difficulty of the detailed deformation approximation, for each branch, in addition to graph-attention-based blocks, we also apply graph pooling and unpooling operations \cite{diehl2019edge} to avoid over-fitting problem and improve the model generalization ability. In conclusion, the approximation via the detail generator after the first layer $W_{\text{detail}}^{[2\text{-}L]}$ can be expressed as: 
\begin{equation}
\|\Delta_{\text{detail}}\|, \delta_{\text{detail}} = W_{\text{detail}}^{[2 \text{-}L]}(\boldsymbol{\mathcal{G}}^{*}).
\end{equation} 
We multiply the predicted $\|\Delta_{\text{detail}}\|$ and $\delta_{\text{detail}}$ to obtain the corrective displacement $\Delta_{\text{detail}}$, and add this to the coarse deformation to obtain the ultimate detailed clothing deformation:
\begin{equation}
M_{\text{detail}} = M_{\text{coarse}} + \Delta_{\text{detail}}.
\end{equation} 
To train the proposed network, we also adopt MSE loss as the loss function to minimize the difference between the predicted $\Delta_{\text{detail}}$ and the ground truth $\Delta_{\text{detail}}^{\text{GT}}$.
\begin{figure*}[t]
     \centering
     \begin{subfigure}[c]{0.32\textwidth}
         \centering
         \includegraphics[width=\textwidth]{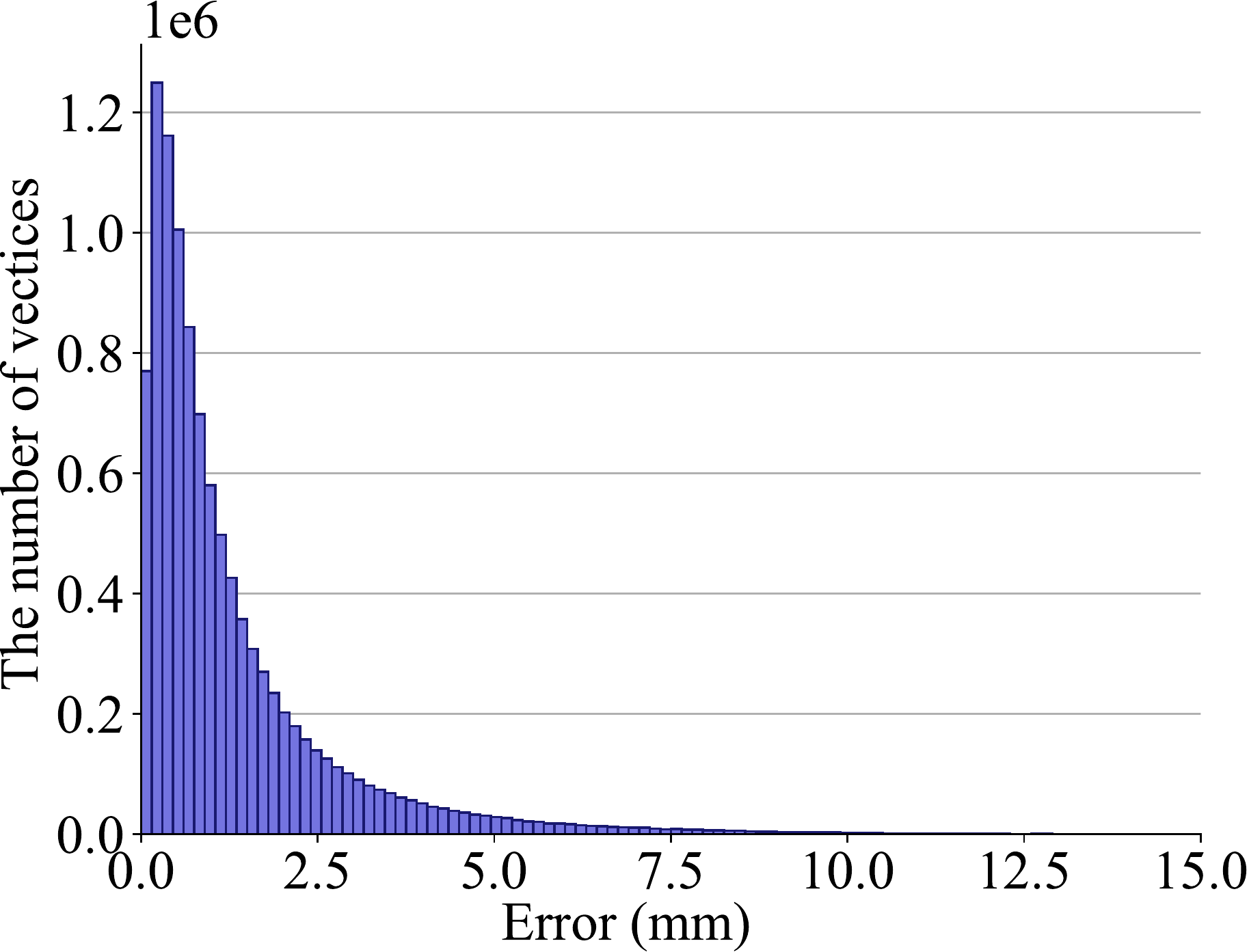}
         \caption{Thin body}
         \label{fig:NewShapeThin_hist}
     \end{subfigure}
     \hfill
     \begin{subfigure}[c]{0.32\textwidth}
         \centering
         \includegraphics[width=\textwidth]{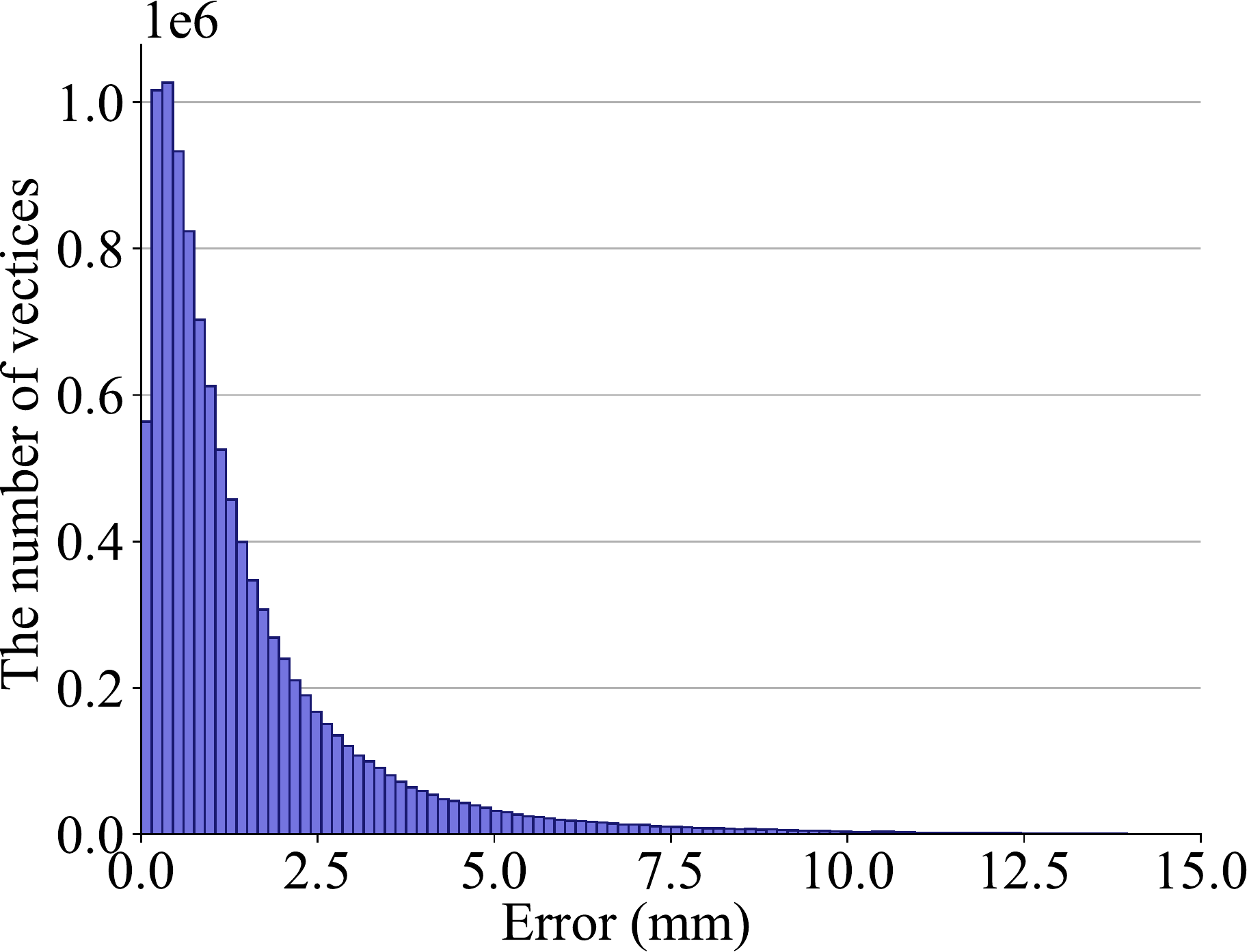}
         \caption{Regular body}
         \label{fig:NewShapeOri_hist}
     \end{subfigure}
     \hfill
     \begin{subfigure}[c]{0.32\textwidth}
         \centering
         \includegraphics[width=\textwidth]{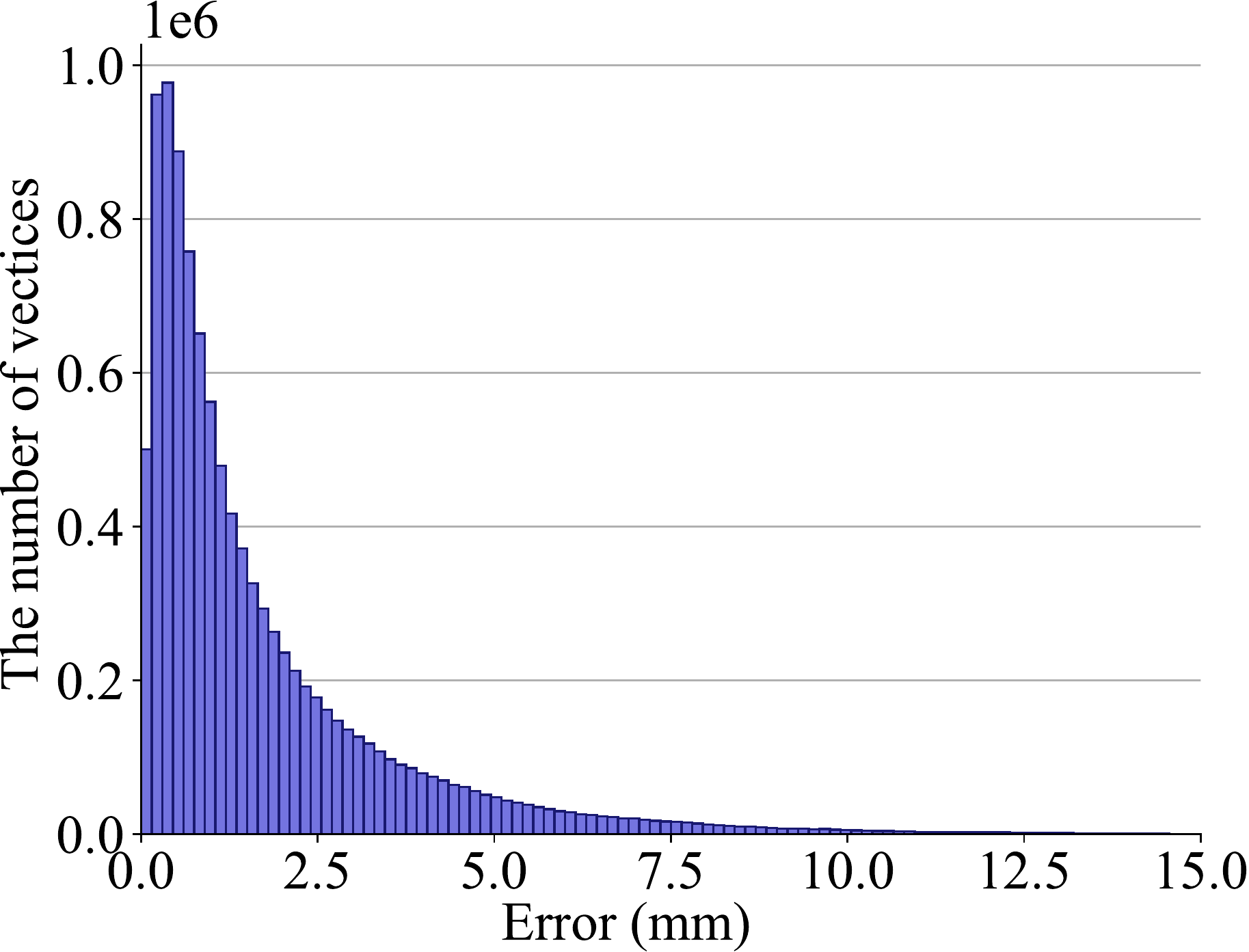}
         \caption{Fat body}
         \label{fig:NewShapeFat_hist}
     \end{subfigure}
        \caption{Histogram plot of distribution of per-vertex errors of generalization to new body shapes. The bin width is 0.15. The first two bars (error: 0-0.15 and 0.15-0.3) in \ref{fig:NewShapeThin_hist} have the largest number of vertices, while those in  \ref{fig:NewShapeFat_hist} have the smallest number of vertices.}
        \label{fig:LogNewShape}
\end{figure*}

\section{Evaluation}
\subsection{Dataset and Implementation}
To evaluate our proposed method, we create a dataset consisting of various garments, body shapes, and animated poses for training and testing. To produce the ground truth data of garments, we utilize the 3D clothing design and simulation software Marvelous Designer \cite{Marvelous} to design and generate deformations for nine t-shirts with different mesh topologies and the number of vertices. To obtain the coarse data, we apply a Laplacian smoothing
operator to each generated clothing mesh. For different body shapes, we select 11 SMPL bodies by sampling the first two shape components. For the pose variation, we choose 3415 animated poses from CMU mocap \cite{CMU} and AMASS dataset \cite{AMASS}, including motion sequences of dancing, ballet, \textit{etc.} In particular, there are six t-shirts and eight bodies forming the 48 garment-body pairs with 3010 poses for optimizing. To verify the effectiveness of the methods, the remaining three garments and three body shapes with 405 poses are used for testing. We randomly select three pairs from the training set and two pairs from the test set as the validation set.   

For the implementation, as shown in Figure \ref{fig:FigPipeline}, we next describe the detailed structure of $W_{\text{coarse}}$, $W_{\text{parser}}$, and $W_{\text{detail}}$. For training $W_{\text{coarse}}$, the features of graph $\boldsymbol{\mathcal{G}}_{b'}$ are input into a GAT block with the hidden feature size of 256 where the multi-head number is 4, the feature sizes of the self-reinforced stream and aggregation stream are 128 and 32 respectively. Features are applied with Tanh activation and then fed into the $\|\Delta_{\text{coarse}}\|$ prediction branch and the $\delta_{\text{coarse}}$ prediction branch, both branches contain three GAT blocks with the hidden feature size of [512, 512, 256]. After graph convolution, three fully connected layers are used to transform the features with the hidden sizes of [256, 128, 1] in the  $\|\Delta_{\text{coarse}}\|$ prediction branch and of [256, 128, 3] in the $\delta_{\text{coarse}}$  prediction branch. To ensure that the output range is reasonable, ReLU and Tanh activation functions are used respectively after each layer of the two branches. Additionally, normalization is also used for features in the $\delta_{\text{coarse}}$ branch. For training $W_{\text{parser}}$, the multi-source attributes $(\alpha, \beta, \theta)$ (where $\alpha \in \mathbb{R}^{3}$, $\beta \in \mathbb{R}^{10}$, $\theta \in \mathbb{R}^{72}$) are transformed into detail-aware encodings by three fully connected layers ([256, 512, 1024]) and ReLU activation function. For training $W_{\text{detail}}$, the graph features $\boldsymbol{\mathcal{G}}$ are fed a GAT block with the hidden feature size of 1024. After infusing graph features with detail-aware encodings, the feature dimension is unchanged and features are input into the $\|\Delta_{\text{detail}}\|$ prediction branch and the $\delta_{\text{detail}}$ prediction branch. The structure of four GAT blocks ([256, 256, 128, 96]) and operations of pooling ($N_g$ roughly becomes half)and unpooling (restored) are the same in each branch. Finally, fully connected layers with the hidden feature sizes of [128, 64, 1] and [128, 64, 3] and corresponding activations are respectively adopted in the $\|\Delta_{\text{detail}}\|$ prediction branch and $\delta_{\text{detail}}$ prediction branch. 


\subsection{Quantitative and Qualitative Evaluation}
\noindent \textbf{Generalization to new bodies.} 
\begin{figure}[t!]
  \centering
  \includegraphics[width= 0.92\linewidth]{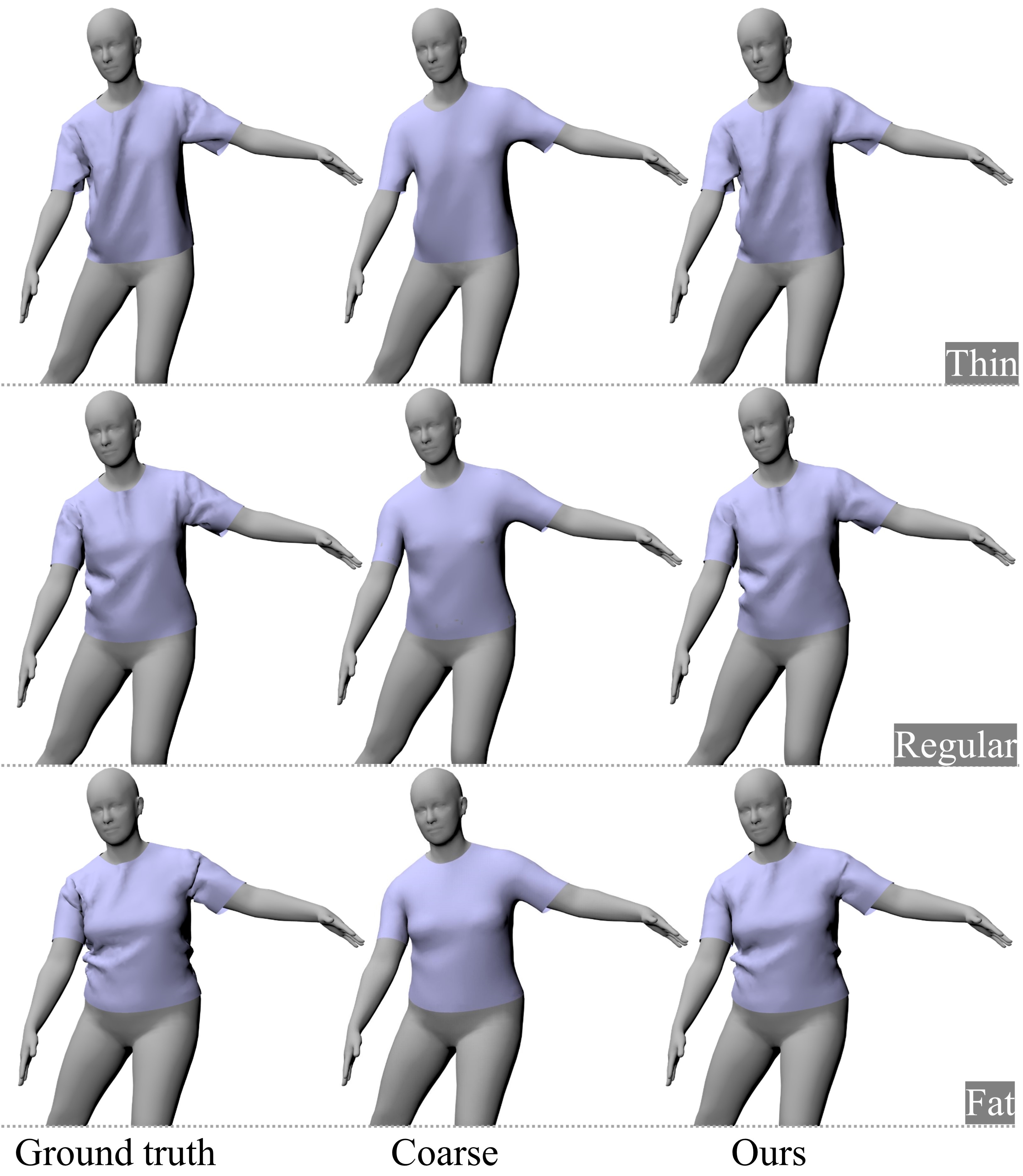}
  \caption{\label{fig:FigGenNewShapes} Generalization to new thin, regular, and fat bodies.
}
\end{figure}
As shown in Figure \ref{fig:FigGenNewShapes}, we provide the generalization results of thin, regular, and fat bodies that are unseen in the training set. Based on the predicted coarse deformation, our method is able to generate fine-scale wrinkles which have no obvious difference with the ground truth data. In addition, our method can successfully predict individualized and detailed clothing deformations of bodies with different shapes, which contains rich and plausible wrinkles in the area of the left side of the waist. During the training, the influencing attributes are transferred into detail-aware encodings and clothing deformation is learned in a detail-aware manner, so that we can effectively make accurate predictions for new body shapes. Quantitatively, in Figure \ref{fig:LogNewShape}, we counted the error distribution of these three test bodies wearing the same training garments under the same training poses. As observed, the number of vertices is the highest in the clothing deformation errors of thin bodies close to zero (Figure \ref{fig:NewShapeThin_hist}), and the mean error of per vertex is about 1.33mm as reported in Table \ref{tab:TabGenNewShapes}. The deformation prediction error of the garment worn by thin bodies is relatively smaller since the clothing folds are simpler than the garment worn by fat bodies; in contrast, the garment worn by fat bodies have more complicated folds, making them relatively difficult to predict. Overall, through the deformation refinement of $W_{\text{detail}}$ and $W_{\text{parser}}$, deformation errors are reduced by about half compared with coarse deformations.

\begin{table}[t]
\begin{center}
\caption{Mean error (mm) of per-vertex deformations in different body shapes.}
 \label{tab:TabGenNewShapes}
\begin{tabular}{c c c c}
 \toprule
Test shapes & Thin & Regular & Fat\\
 \midrule
 Coarse  & 2.82 & 3.01 & 3.27 \\
 Detail & 1.33 & 1.52 & 1.74 \\
 \bottomrule
\end{tabular}
\end{center}
\end{table}

\begin{figure*}[t]
     \centering
     \begin{subfigure}[b]{0.32\textwidth}
         \centering
         \includegraphics[width=\textwidth]{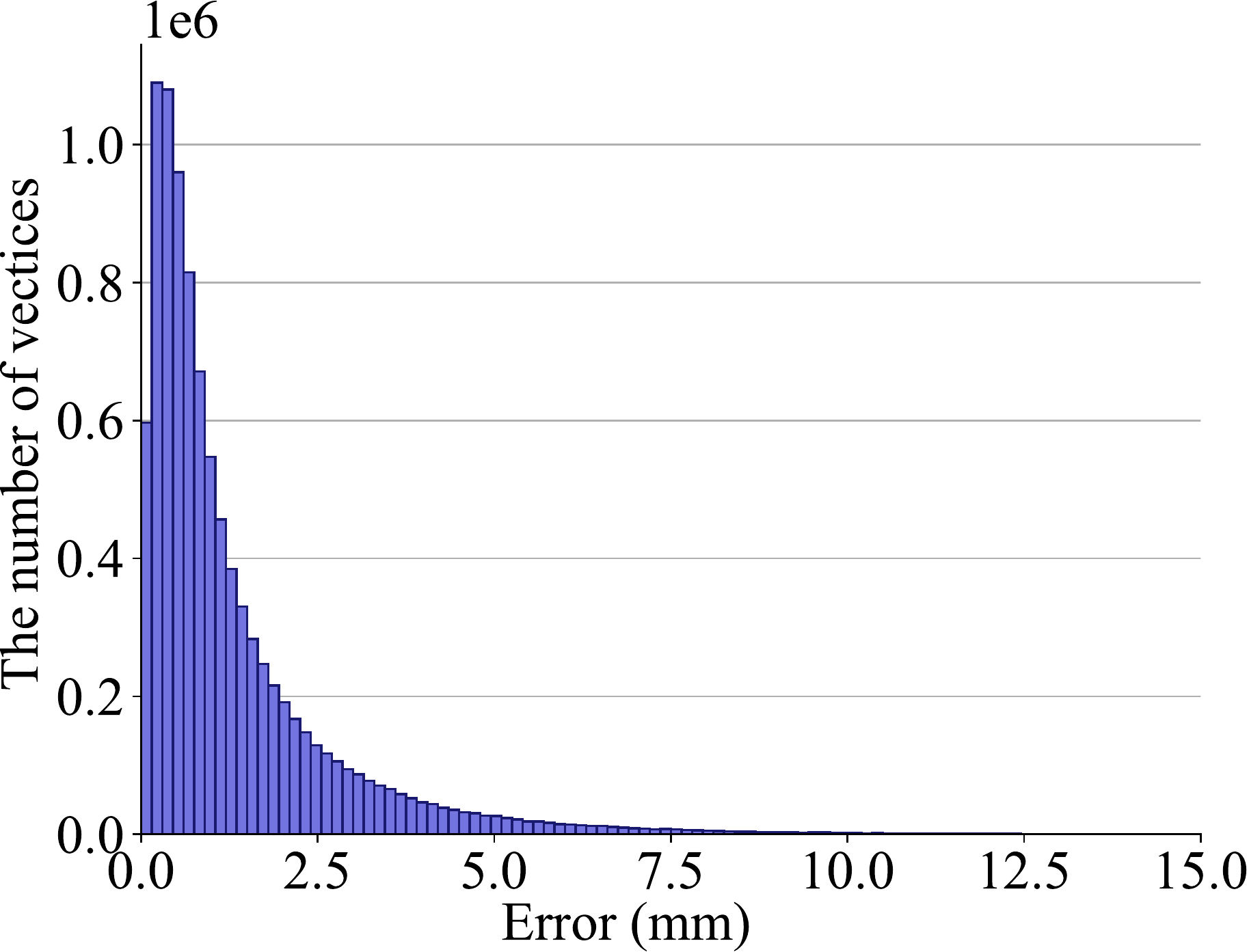}
         \caption{Short garment}
         \label{fig:NewGarShort_hist}
     \end{subfigure}
     \hfill
     \begin{subfigure}[b]{0.32\textwidth}
         \centering
         \includegraphics[width=\textwidth]{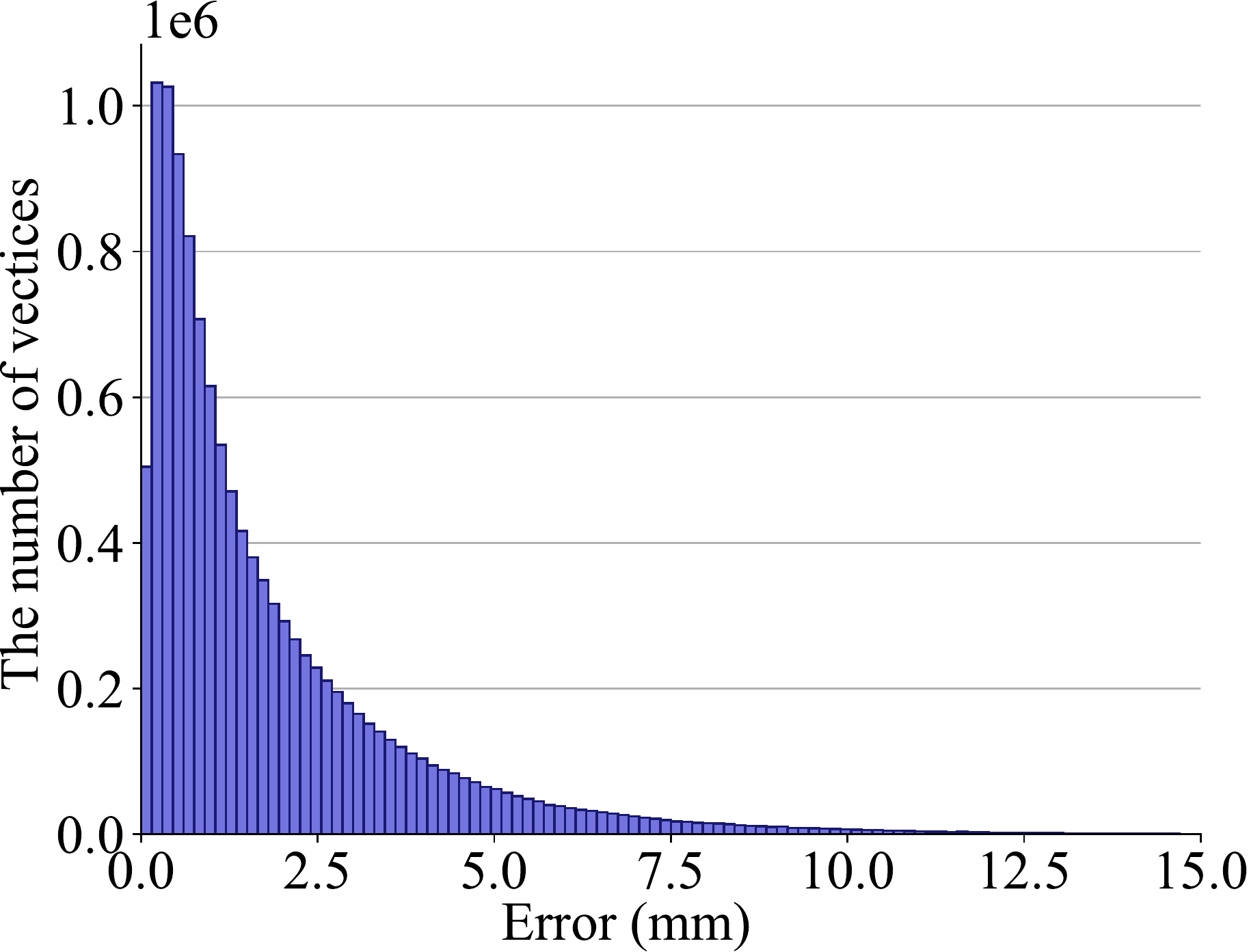}
         \caption{Long garment}
         \label{fig:NewGarLong_hist}
     \end{subfigure}
     \hfill
     \begin{subfigure}[b]{0.32\textwidth}
         \centering
         \includegraphics[width=\textwidth]{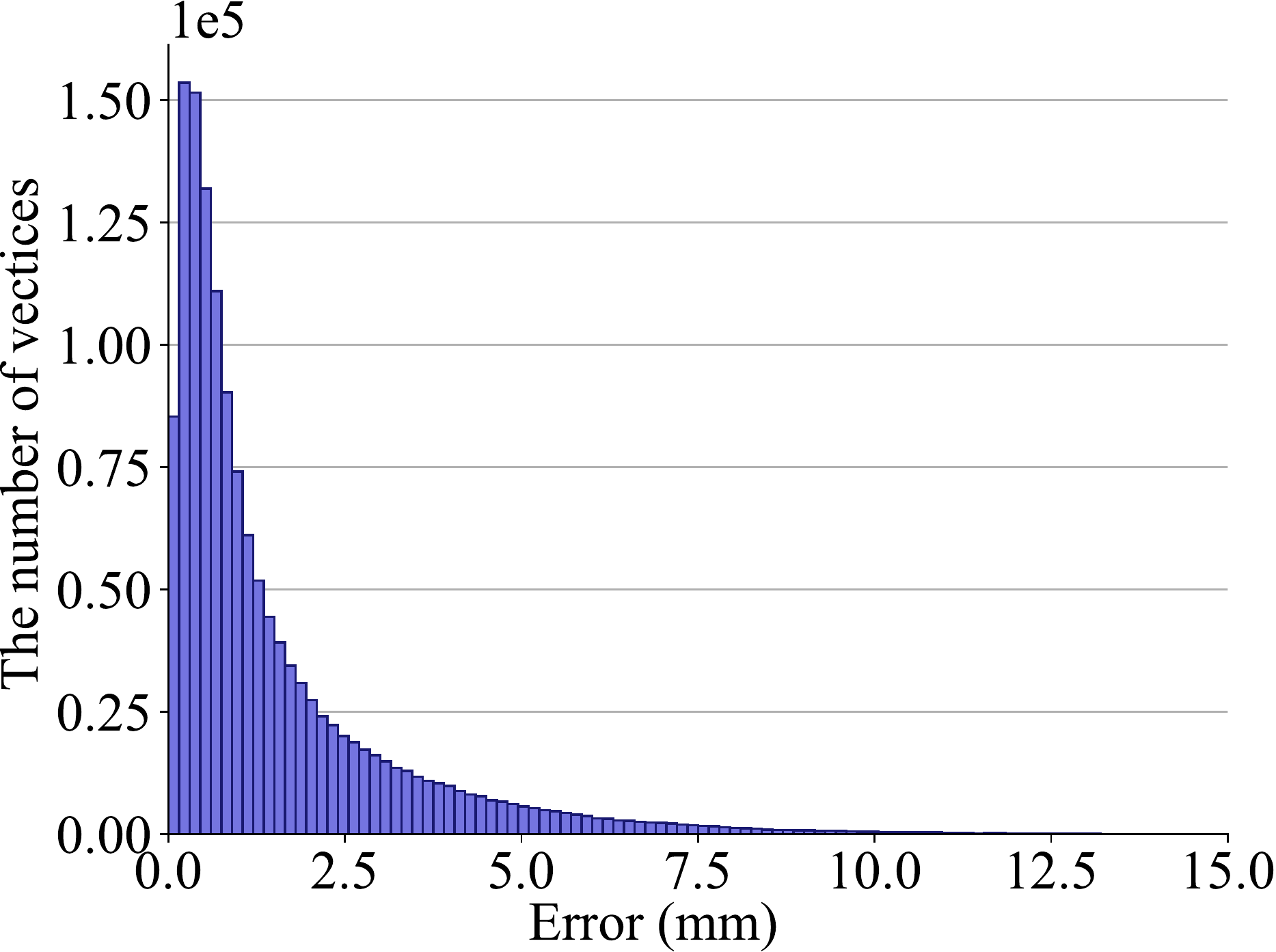}
         \caption{New poses}
         \label{fig:NewPose_hist}
     \end{subfigure}
        \caption{Histogram plot of the distribution of per-vertex errors of generalization to new garments and new poses. The bin width is 0.15.}
        \label{fig:three graphs}
\end{figure*}

\begin{figure}[t!]
  \centering
  \includegraphics[width= 0.9\linewidth]{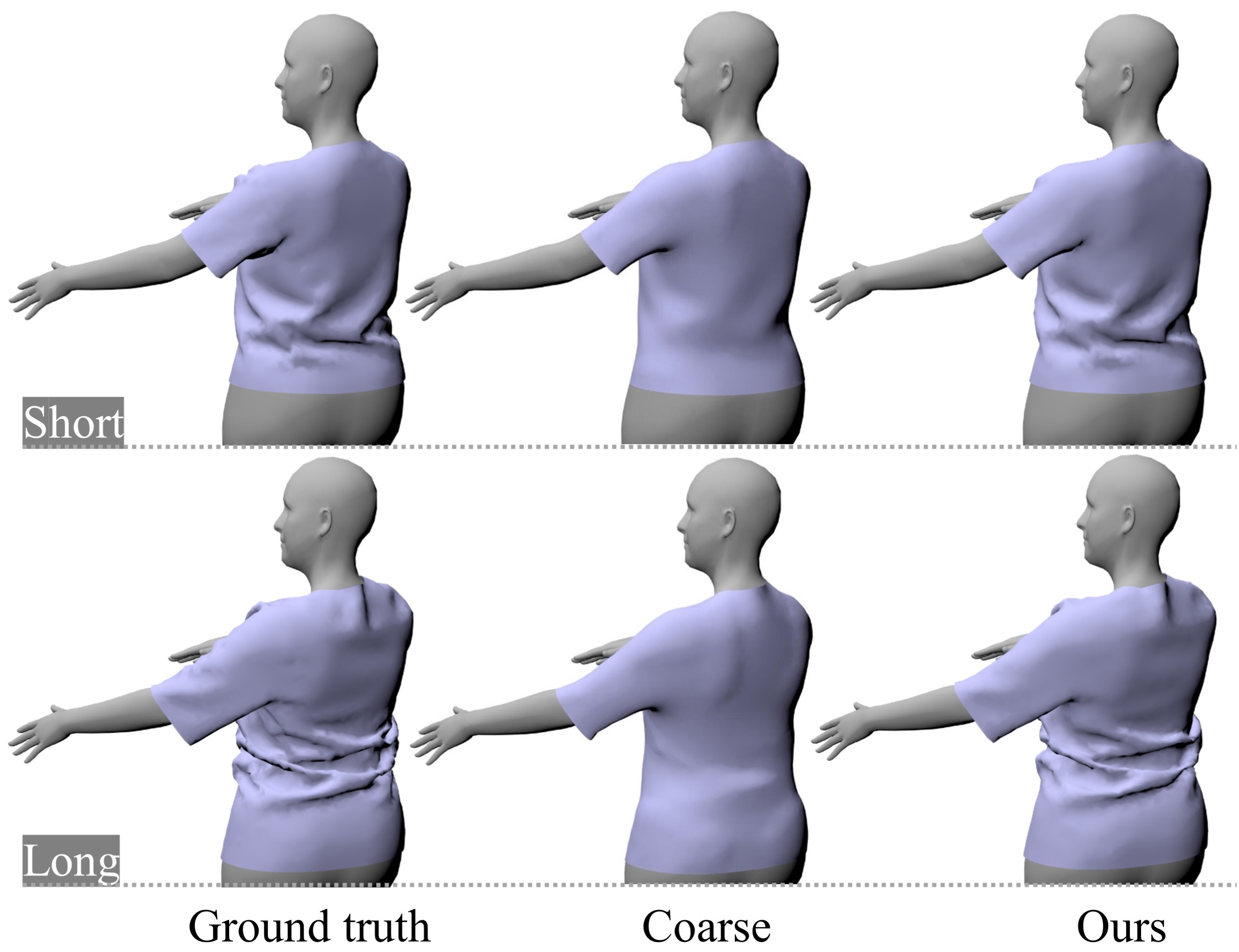}
  \caption{\label{fig:FigGenNewGar} Generalization to new short and long garments.
}
\end{figure}

\noindent \textbf{Generalization to new garments.} Figure \ref{fig:FigGenNewGar} shows the qualitative results of the generalization to new garments with short and long lengths. Here, the tested garments have different mesh topologies and different number of vertices from the training. Thanks to the graph-learning-based model and proposed detail-aware strategies, our model can reasonably approximate deformations with rich details regardless of the mesh topology. From the perspective of fit, long garments fit tightly to the body due to the influence of the hem, so there are more denser wrinkles in the deformation result, which also follows the law of our first observation as stated in Section \ref{sec:fit}. Figure \ref{fig:NewGarShort_hist} and \ref{fig:NewGarLong_hist} show the distribution of vertex errors for these test garments, worn by the same training bodies in the same poses. Due to the rotation of the upper body, the folds are mostly concentrated around the waist. For long garments with the tighter fit, the hem of the garment is stuck in the hip, which causes more folds at the waist to accumulate, resulting in relatively larger errors as corresponding to the results in Table \ref{tab:TabGenNewGar}.

\begin{table}[t!]
\begin{center}
\caption{Mean error (mm) of per vertex of deformations in different garments.}
 \label{tab:TabGenNewGar}
\begin{tabular}{c c c}
 \toprule
Test garments & Short & Long\\
 \midrule
 Coarse   & 3.21 & 3.93 \\
 Detail  & 1.38 & 1.77 \\
 \bottomrule
\end{tabular}
\end{center}
\end{table} 
  
\begin{figure}[t!]
  \centering
  \includegraphics[width= 0.8\linewidth]{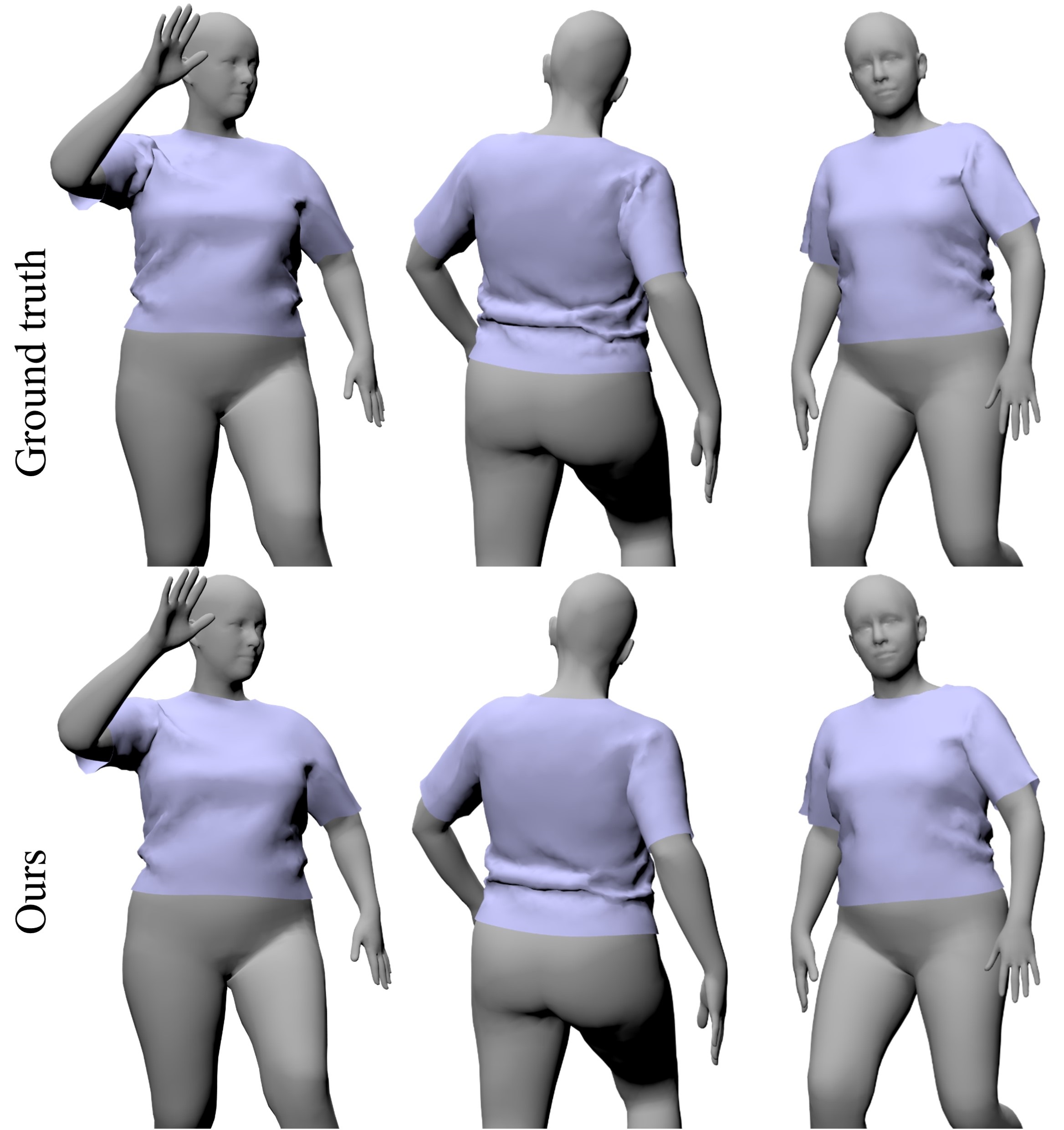}
  \caption{\label{fig:FigGenNewPose} Generalization to new poses.
}
\end{figure}

\begin{figure}[t!]
  \centering
  \includegraphics[width= 0.67\linewidth]{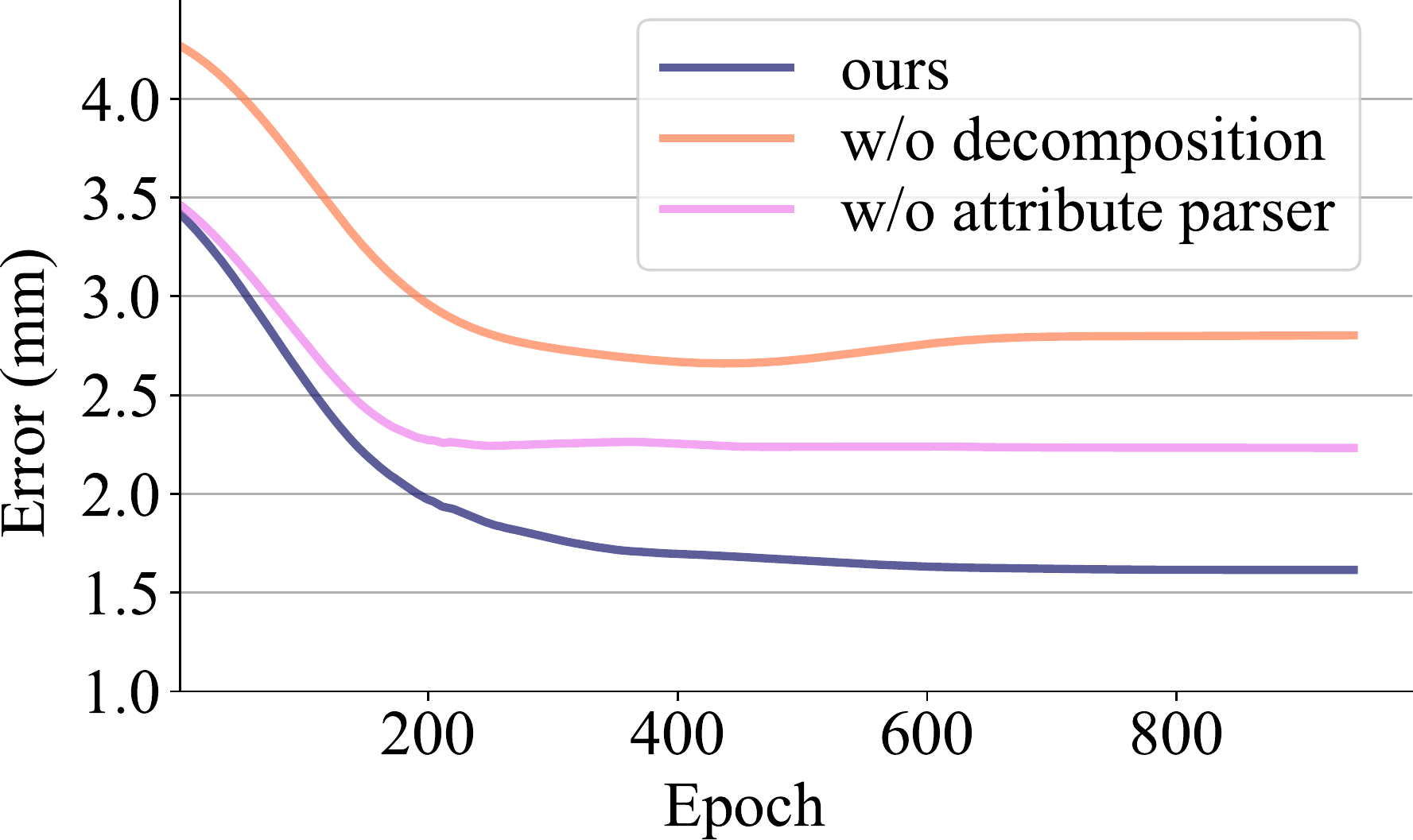}
  \caption{\label{fig:FigLoss} Mean vertex error during validation of generating detail deformations. Our proposed output decomposition and attribute parser play a key role in the learning of detailed deformation.
}
\end{figure}

\begin{figure*}[t!]
\includegraphics[width=0.96\textwidth]{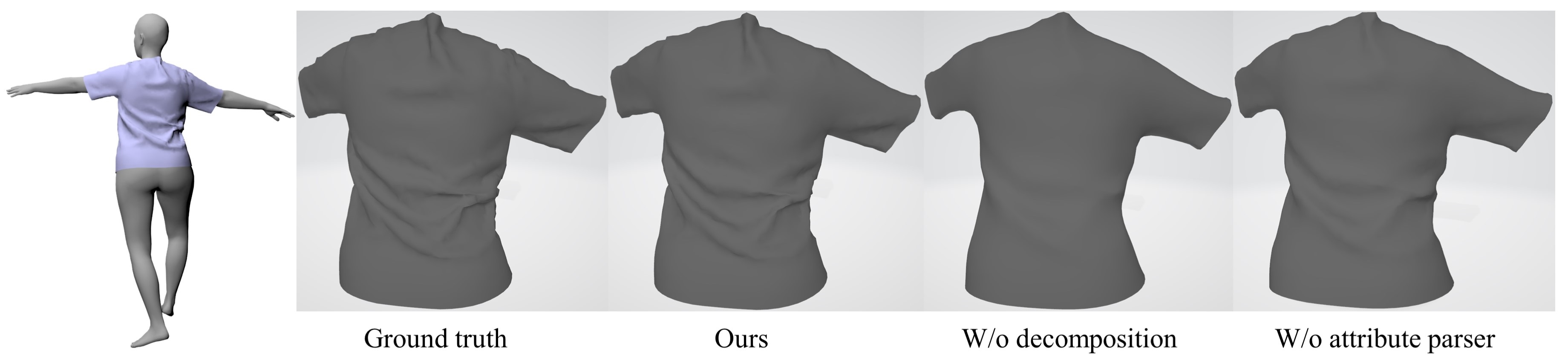}
\centering
\caption{Qualitative results of ablation study comparing detailed deformations approximated by our full method, our method without output decomposition, our method without detail-aware attribute parser, and ground truth of physics-based simulation.
}  
\label{fig:FigAbla}
\end{figure*}
 
\noindent \textbf{Generalization to new poses.}  Figure \ref{fig:FigGenNewPose} shows the results of visually evaluating the quality of our proposed approach of generalization to new poses, in which we compared the deformations of the ground truth physics-based simulations and our predictions. We animated dressed bodies with new postures of raising the hand, walking, and swinging. Through the proposed method, attractive details can be successfully generated, in which the wrinkles in areas of armpits, waist, shoulders are rich and quite similar to real effect of the ground truth. During the training, in addition to graph constructions, we also design a $W_{\text{parser}}$ to generate detail-aware encodings and infuse them into the graph neural network, so that the model can learn the individualized deformations caused by different poses. Figure \ref{fig:NewPose_hist} shows the distribution of errors for the training garment and body with test motions. As observed, the number of vertices decreases roughly exponentially with the distance error, which demonstrates the good generalization ability of our proposed method for new poses.   

\subsection{Ablation Study}

We conducted an ablation study to highlight the effectiveness of the proposed strategies: output decomposition and detail-aware attribute parser. Specifically, we trained the detail generator model by individually retaining the output displacement of per-vertex as the original three-dimensional vector $\Delta_{\text{detail}}$, and removing the attribute parser $W_{\text{parser}}$ where attributes are directly assigned to each graph node. In Figure \ref{fig:FigLoss}, we plot the mean error of per-vertex during the validation process. In the beginning, the orange line (without output decomposition) has the largest error, because the output result is difficult to be approximated with three values from negative infinity to positive infinity. As the epoch increases, it is still accompanied by highly complicated outputs and errors remaining between 2.5-3mm that cannot be reduced. For the pink line (without the detail-aware attribute parser), it can keep dropping for the first 200 epochs and achieves relatively better accuracy than the method without decomposition because the output value range is narrowed, which can effectively ensure that the prediction is reasonable to a certain extent. However, the error after stabilization is still not ideal. In contrast, with the help of the proposed strategies, our method can well generalize to unseen objects and achieve minimum average error in predictions. The corresponding qualitative evaluation is depicted in Figure \ref{fig:FigAbla}. Ground truth data with try-on effect is shown in the leftmost. We also enlarge the garment alone, which contains various details on the collar, back waists and sleeves. The resulting deformation of keeping output as the original $\Delta_{\text{detail}}$ has the smoothest appearance, because it suffers the extremely complicated challenge of using one model to process information from a variety of garments, body shapes, and poses and requiring numerical inferences within an infinite range. For the results of without attribute parser, the deformation has the obvious folds with the wider width in the waist and collar areas, which can reflect a certain degree of wrinkle trend. However, by directly assigning unprocessed attributes to every single node, only $W_{\text{detail}}$ cannot produce personalized fine-scale details based on the given instance (attributes and features). In contrast, our full method is able to successfully generate these major and subtle wrinkles and recover detail effects similar to the ground truth, thanks to the proposed strategies of output decomposition and attribute parser.

\subsection{Comparison}
\begin{figure*}[t]
\includegraphics[width=0.9\textwidth]{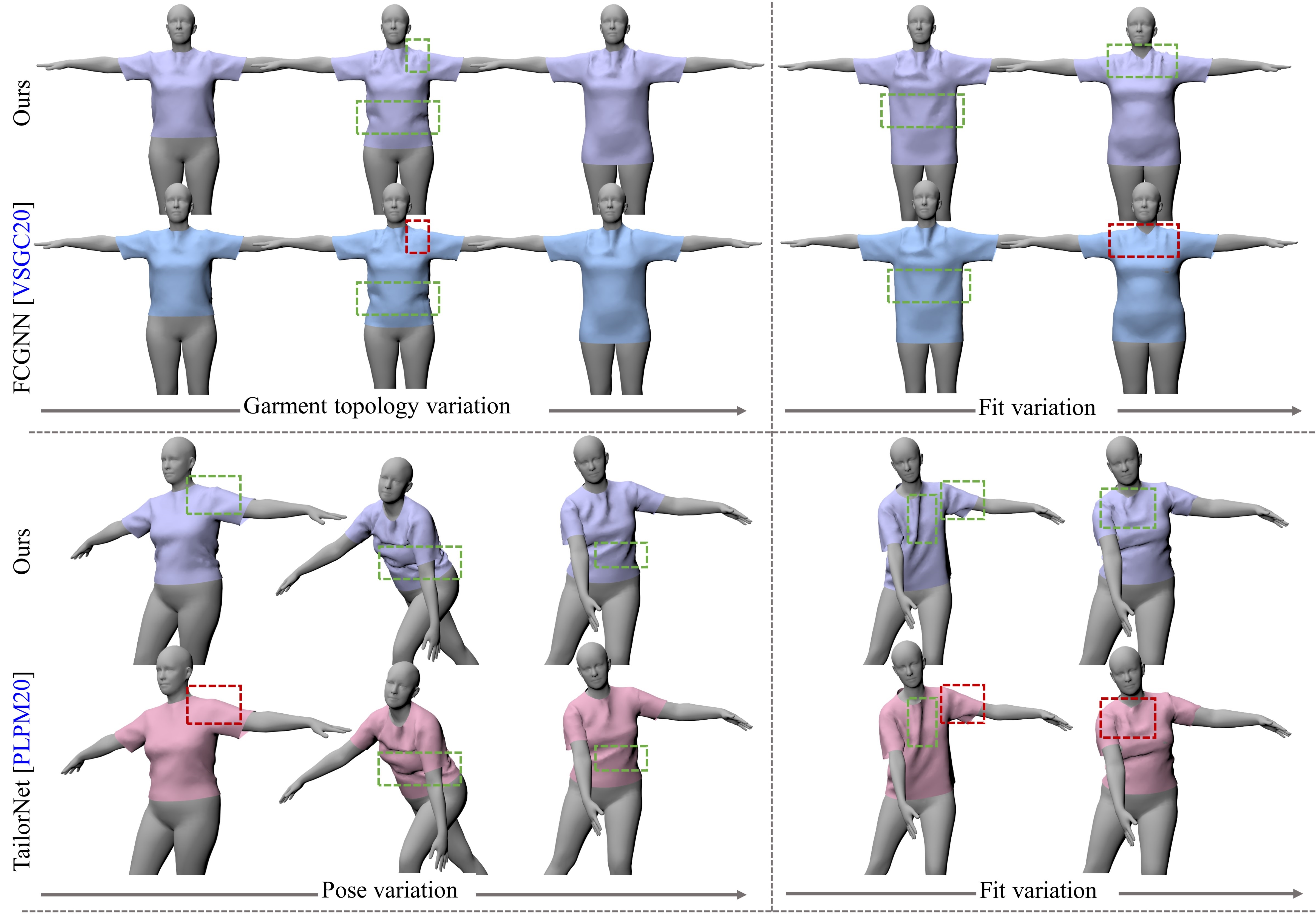}
\centering
\caption{Qualitative comparison of  FCGNN\cite{vidaurre2020virtualtryon}, TailorNet \cite{patel20tailornet}, and our method. FCGNN (with blue garments) can achieve mesh topology variation and fit variation; TailorNet can achieve pose variation and fit variation; Ours is the first approach to achieve all of these. In addition, our method can generate detail-aware clothing animation, that allows for rich detail prediction caused by various attributes.    
}  
\label{fig:FigCompare}
\end{figure*}

We compare our method with other state-of-art learning-based approaches: FCGNN \cite{vidaurre2020virtualtryon} and TailorNet \cite{patel20tailornet}. As listed in Table \ref{tab:DiffMethodTab}, FCGNN can generalize to arbitrary garment meshes due to the use of a fully convolutional graph neural network, but it only predicts clothing deformations under the t-pose. TailorNet is able to achieve pose-dependent deformations, but it is limited to the use of trained MLP models to predict deformations of new mesh topologies. To the best of our knowledge, currently, there is no prior research involving tasks that are exactly the same as our method to approximate clothing deformations for various mesh topologies and body shapes in diverse poses. 

\begin{table}[t!]
\begin{center}
\caption{Comparison of our approach with other state-of-art learning-based clothing deformation methods. Our method can achieve more functions with smaller model size.}
 \label{tab:DiffMethodTab}
\begin{tabular}{c c c c}
  \toprule
Methods  &  \makecell[c]{Topology \\variation} & \makecell[c]{Pose \\variation} &  \makecell[c]{Fit \\variation} \\
 \midrule
 FCGNN & \cmark  &   \xmark &  \cmark \\
TailorNet  &  \xmark  &   \cmark &  \cmark  \\
 Ours &  \cmark  &   \cmark &  \cmark \\
 \bottomrule
\end{tabular}
\end{center}
\end{table}

Figure \ref{fig:FigCompare} shows a qualitative comparison of our method and other two approaches. Because of the limited terms listed Table \ref{tab:DiffMethodTab}, garments with different mesh topologies (the number of vertices: $N_g = 3178, 3311, 3666$) and garments worn by different body shapes are evaluated for the method of FCGNN \cite{vidaurre2020virtualtryon}; one garment under different postures and worn by different body shapes are evaluated for the method of TailorNet \cite{patel20tailornet}. In addition, the part of models and dataset are not public yet, so in this experiment of achieving the detailed deformation, we use our dataset (the coarse deformation and the ground truth) and set features and networks according to the implementation mentioned in the respective methods. As observed, both FCGNN and TailorNet have the generalization abilities and are capable of generating plausible deformation effects, especially in waist areas, and for garments worn by thin body shapes (as green-framed parts in the figure). Despite the predictions for conspicuous wrinkles, the shoulder areas with small fine-scale wrinkles are still overly smooth (as red-framed parts in the figure). The comparison shows the benefit of our proposed method: even in the face of multi variations, the model still has excellent generalization ability to approximate not only obvious wrinkle folds but also fine-scale details.  

\subsection{Runtime Performance and Memory}
With the nVIDIA GeForce RTX2080Ti GPU, in the approximation of clothing deformations of meshes with 3000-4000 number of vertices, the average per-frame runtime is about 21ms, where coarse prediction takes 8ms and detail prediction takes 13ms. The proposed method is 50 times faster than physics-based simulation, making it suitable for real-time applications. The memory footprint of our method is about 37.4MB, where the $W_{\text{coarse}}$ model is 18.4MB, and the $W_{\text{detail}}+ W_{\text{parser}}$ model is 19MB.

\section{Conclusion}
\begin{figure}[t!]
  \centering
  \includegraphics[width= 0.96\linewidth]{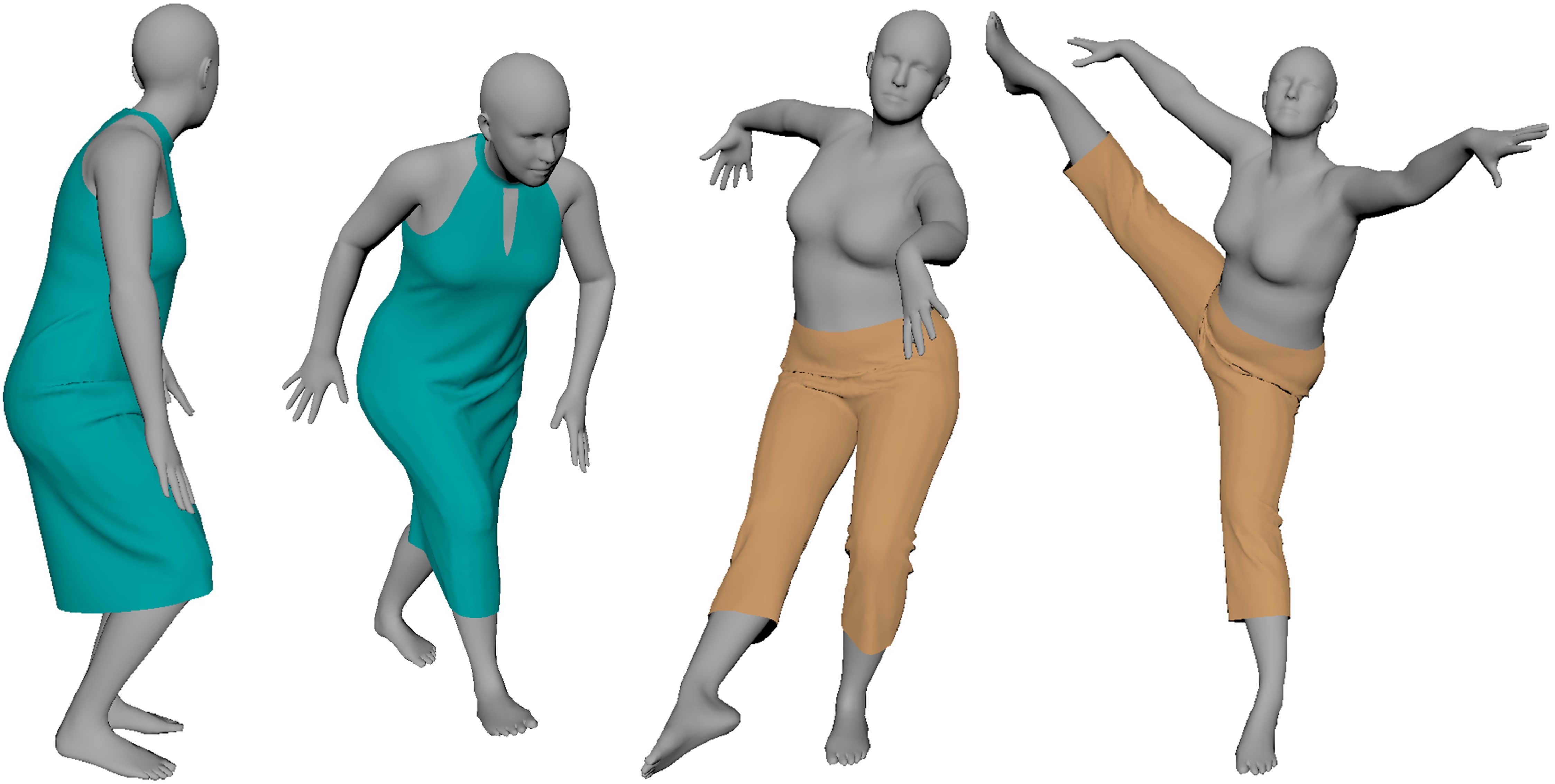}
  \caption{\label{fig:FigPantsDress} Our method can also make predictions for dresses and trousers under unseen poses. Realistic deformations containing fine-scale wrinkles can be approximated. 
}
\end{figure}
We have presented a graph-learning-based deformation method for garments whose mesh topology can be arbitrary and can be worn by any body shape in various poses. To achieve generalization and high-quality predictions at the same time, we first propose the fit parameter as one of the important attributes influencing the wrinkle details. Then, we design an attribute parser to generate detail-aware encodings and infuse them into the graph neural network to help generate individualized details. Last and most importantly, we propose a novel output reconstruction strategy for the excellent convergence of extremely complex regressions. This strategy can not only be adopted in clothing deformations, but also works for predicting positions or displacement adjustments in other areas. Experimental results have shown that our method with the above technical innovations can overcome the limitations and outperforms existing learning-based approaches. 

Despite achieving powerful generalization and impressive detailed deformations, our method still has a few drawbacks that can be addressed in future works. First, we currently use a postprocessing step to deal with garment-body interpenetrations. Since each garment-body pair in the dataset is selected in advance, collision is not a serious problem and can be effectively fixed by postprocessing. In the future, we will investigate a loss term (\textit{e.g. }, the self-supervised term \cite{santesteban2021self}) for our method to penalize the interpenetration at training time. Secondly, in addition to deforming t-shirts, our method can also make deformation predictions for dresses and trousers (shown in Figure \ref{fig:FigPantsDress}). For each category of garments, we train the network separately. In the future, it would be interesting and practical to explore a super unified model that can generate deformations for any garment type while retaining details. Thirdly, each type of clothing in our dataset is given the same material settings and we do not consider material-dependent deformations in our method. Future work can expand the dataset to include various materials, and then explore related material attributes as network inputs to automatically generate more realistic deformations.

{\small
\bibliographystyle{splncs04.bst}
\bibliography{egbib.bib}
}
\end{document}